\documentclass{article}
\usepackage{spconf,amsmath,graphicx, amssymb, xcolor}
\usepackage{multirow}
\usepackage{hyperref}

\title{Transducer Consistency Regularization for Speech to Text Applications}

\name{Cindy Tseng, Yun Tang, Vijendra Raj Apsingekar}
\address{Samsung Research America, USA}

\graphicspath{ {./images} }

\begin{document}
%
\maketitle
\begin{abstract}

Consistency regularization is a commonly used practice to encourage the model to generate consistent representation from distorted input features and improve model generalization. 
It shows significant improvement on various speech applications that are optimized with cross entropy criterion. However, it is not straightforward to apply consistency regularization for the transducer-based approaches, 
which are widely adopted for speech applications due to the competitive performance and streaming characteristic. The main challenge is from the vast alignment space of the transducer optimization criterion and not all the alignments within the space contribute to the model optimization equally. 
In this study, we present Transducer Consistency Regularization (TCR), a consistency regularization method for transducer models. We apply distortions such as spec augmentation and dropout to create different data views and minimize the distribution difference. We utilize occupational probabilities to give different weights on transducer output distributions, thus only alignments close to oracle alignments would contribute to the model learning.  
Our experiments show that the proposed method is superior to other consistency regularization implementations and could effectively reduce the word error rate (WER) by an average of 3.56\% relative to a strong baseline on the \textsc{Librispeech} dataset.
\end{abstract}
\begin{keywords}
Consistency Regularization, Transducer, Automatic Speech Recognition, Speech Translation
\end{keywords}
\section{Introduction}
\label{sec:intro}

Transducer model was first introduced in~\cite{graves2012sequence} to transform any input sequence into another finite, discrete output sequence.  It has been widely adopted in automatic speech recognition (ASR)~\cite{Zhang2020TransformerTA,yeh2019transformertransducer,Li2022RecentAI} due to its excellent performance~\cite{chiu2019comparison,Li2020OnTC} and good support for both streaming and offline decoding modes~\cite{Wu2020StreamingTA,Shi2020EmformerEM}. Recently, researchers also showed that the transducer model could achieve very competitive results in other challenge tasks, such as speech to text translation (ST)~\cite{Liu2021CrossAA,Xue2022LargeScaleSE,Tang2023HybridTA}, which is a non-monotonic alignment task, and text to speech synthesize (TTS)~\cite{Kim2023TransduceAS}, which is with a longer target sequence than the input sequence.  

Consistency regularization is an approach to learn to generate consistent representations of the input feature with different views~\cite{Chen2020ASF,Liang2021RDropRD} or from different models~\cite{grill2020bootstrap} by minimizing the difference in distribution. It is widely used in self-supervised learning and semi-supervised learning to stabilize training and alleviate data perturbation sensitivity 
~\cite{Miyato2017VirtualAT,Sato2019EffectiveAR,Chen2020ASF,Liang2021RDropRD}.
It is a desirable property for speech to text tasks since speech data contain different kinds of variations, such as channel variability, speaker variability, and background noise etc. ~\cite{Min2022SANAR} propose to learn the consistency among two sub-networks for the Chinese ASR task;  
~\cite{gao2023empirical} apply consistency regularization for distributions with different input modalities or same input modality with different dropout operations for the ST task. Both studies show consistency learning is beneficial for the speech to text tasks,  though their studies are limited on the encoder-decoder framework optimized with cross entropy loss.  

Applying consistency regularization for the transducer-based modeling is not straightforward due to the complexity of the output distributions. In transducer optimization~\cite{graves2012sequence}, the model would consider all potential alignments among input and output sequences. Not all alignments are practical given the input and output sequences and the distributions from those alignments could far from the desirable distributions we expected. 
As shown in ~\autoref{tab:librispeech_results}, consistency regularization on those distributions could hurt the model performance.

In this work, we focus on applying consistency regularization during the transducer optimization. We create different views of model outputs based on different data augmentations and dropout operations. The model output distributions are treated differently based on their contributions to the optimization loss during training.  
We assign different weights to different distributions based on their contributions to the overall transducer loss. 
Distributions that are close to the oracle alignment would be assigned more weights, while distributions that are far from the oracle alignment would contribute less to consistency regularization. 
We compare our proposed method with other consistency regularization techniques on the 
\textsc{Librispeech}~\cite{panayotov2015librispeech} data set. The results show that our proposed method is superior compared to other implementations and could reduce 0.2 WER on top of a strong baseline.  

Our contributions are summarized as follows:
\begin{itemize}
  \item We propose to generate different views of the input data by different data augmentation and dropout operations in the consistency learning 
  \item We propose an approach to assign different weights to output distributions when applying the consistency regularization on transducer-based models
  \item Our method achieves the best results compared with other consistency regularization implementations and obtained 0.2 WER reduction compared with a strong baseline 
\end{itemize} 

\begin{figure}[t]
    \centering
    \includegraphics[width=.8\columnwidth]{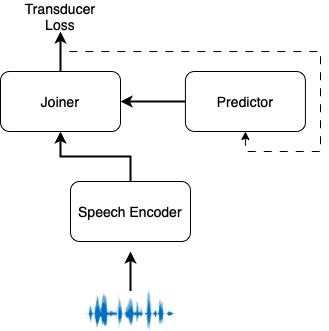}
    \caption{Transducer model. Composed of encoder, predictor, and joiner. The encoder transforms current audio data into hidden audio representation, predictor receives the previous token and predict the hidden text representation, and joiner takes the two hidden representations to predicts the current token.}\label{fig:transducer}
\end{figure}
\begin{figure}[t]
    \centering
    \includegraphics[width=0.8\columnwidth]{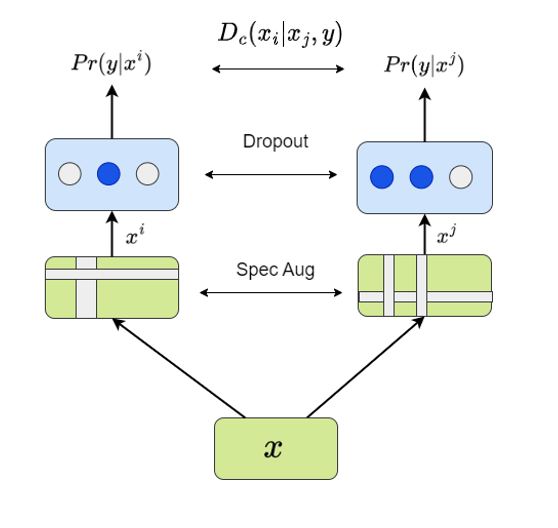}
    \caption{TCR flow chart. The data is first duplicated, then distorted with spec augmentation and dropout to create two different views of data representation. Finally KL divergence is used to minimize the representation difference.}\label{fig:tcr}
\end{figure}

\section{Related Work}
\label{sec:relatedwork}

Consistency regularization is a popular method in self-supervised and semi-supervised learning to learn consistent representation from different data views. Noise injection, cropping, spec augmentation, back translation\cite{xie2020unsupervised} etc are common techniques to create different data views. The model is tasked with finding a common data representation among the different data views. 

\cite{grill2020bootstrap} is a popular semi-supervised learning technique introduced in computer vision. From a given target representation, it trains enhanced online representations. The target model has the same architecture as that of the online architecture, but uses different weights. The target network uses exponential moving average parameters from the online network. Data representation is learned by minimizing the mean squared error between the target representation and the online representation. This would not apply in our case since our final representation is from the joint output probability distribution. The mean squared error is not a good distance metric for probability distributions.

\cite{sapru2022using} applies consistency regularization on the transducer model with semi-supervised learning. It requires two stages training: First stage to pre-train the encoder and the second stage for end-to-end training. The unlabeled data are either weakly augmented or strongly augmented. It reconciles the differences by minimizing the encoder output only, which is trained with cross entropy loss. In our work we don't require a two stage training and we compare the representation directly at the joint output. 

\cite{xie2020unsupervised} focuses on data augmentation techniques applied in semi-supervised training. \cite{masumura2020sequence} applies consistency regularization on ASR tasks. Both works minimize the representational difference with models trained with cross entropy loss. This would not apply in our model, which is trained with transducer loss. 

\cite{Liang2021RDropRD} \cite{gao2023empirical} introduces consistency regularization for models trained in cross entropy loss in supervised learning. They created two different data views by applying different dropouts to the same data. Then they run the model twice with these data to obtain two different model outputs, or duplicate the data in each training and effectively double the batch size. Finally, KL Divergence was applied to minimize the distribution differences between the two outputs. This has shown a great improvement in both ASR and NMT applications. In our work besides of dropout we also add spec augmentation to create different views. Their method only works on models trained with cross entropy loss and cannot be applied to our model which is trained with transducer loss. 

\cite{yu2020dual} and \cite{panchapagesan2021efficient} both apply knowledge distillation on transducer models. They argue that using the entire joint output is memory and compute intensive. Hence, they distill the information by compressing the vocabulary space into three categories, valid token, a special blank token (see ~\autoref{sec:prelim_transducer}), and remaining tokens. However, in both studies, knowledge distillation was between a teacher to a student; either from a bigger model to a small model or from a full context to a streaming context. None of the scenarios are applicable to our case. 

\cite{kuang2022pruned} aims to optimize transducer training that is faster and more memory efficient. It achieves this by first using a trivial joiner that is linear to the encoder and the decoder output to select the pruning bounds. Then it uses the full joint on the selected pruning bounds for the final evaluation. This work focuses mainly on improving the overall efficiency of transducer loss training. But we have not seen this work extended to other domains such as consistency regularization.

\section{Preliminary}
\label{sec:prelim_transducer}
\begin{figure}
    \centering
    \includegraphics[width=.9\columnwidth]{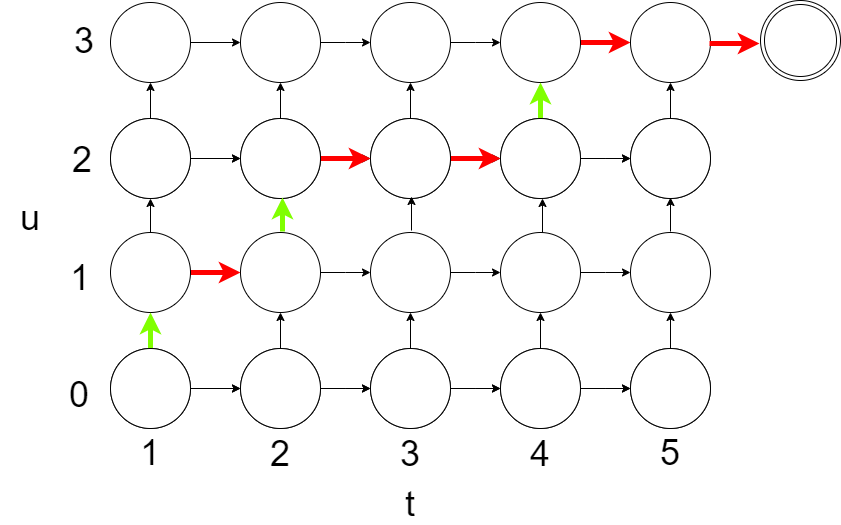}
    \caption{Transducer decoding lattice. Red horizontal arrow represents blank token emission, while green vertical arrow represents non blank token emission. One full path starts from $(t, u) = (1, 0)$ and ends at $(t, u) = (T, U)$.}\label{fig:lattice}
\end{figure}

Transducer model is depicted in \autoref{fig:transducer}. It consists of three main components, the encoder, which maps audio frames to acoustic representations. The predictor, which uses past tokens to predict the next non-blank token. And finally, the joiner, which combines both encoder and predictor outputs to produce a log probability distribution.  

Let $\mathbf{x}_{1:T} = (x_1, x_2, \cdots, x_T)$ be a length $T$ input sequence and $\mathbf{y}_{1:U} =  (y_1, y_2, \cdots, y_U)$ denote the output target sequence. %
Transducer model monotonically maps $\mathbf{x}_{1:T}$ to $\mathbf{y}_{0:U}$, where $y_0 = \langle bos \rangle$ is the begin of sentence token. 
We refer to such mappings $\mathbf{a}=[(t_i, u_i)]_{1\le i \le U+T}$ between $\mathbf{x}_{1:T}$ and $\mathbf{y}_{0:U}$ as {\it alignments}, where $t_i \in [1, T]$ and $u_i \in [0, U] $.   
Transducer model is based on the frame synchronous based decoding~\cite{Dong2020ACO}, i.e., every $x_t$ will generate one or multiple tokens. 
Each green \textcolor{green}{ $\uparrow$ } in the highlighted alignment in ~\autoref{fig:lattice} denotes a new token generation. If no token is generated at $x_t$, a blank token $\varnothing$ is assigned. For example, there are two consecutive \textcolor{red}{ $\rightarrow$ } from step 2 to step 4, and a blank token $\varnothing$ is assigned to $x_3$ in ~\autoref{fig:lattice}.

Transducer minimizes condition probability $Pr(\mathbf{y}|\mathbf{x})$ by marginalizing all possible alignment paths and could be done by the forward-backward algorithm~\cite{graves2012sequence}. The forward variable $\alpha(t,u)$ is the probability of generating
$y_{1:u}$ at step $t$ from the beginning.
\begin{eqnarray}
   \alpha(t,u) & = & \alpha(t-1, u)\varnothing(t-1, u) \nonumber\\
   &  & + \alpha(t,u-1)y(t,u-1) \label{equ:forward}
\end{eqnarray}
where $\alpha(1,0)$ = 1, $y(t,u)=Pr(y_{u+1}|t,u)$ is the emission probability from token $u$ at time $t$ to generate token $u+1$ (vertical movement in the lattice from ~\autoref{fig:lattice});
 $\varnothing(t,u)=Pr(y_u|t, u)$ is corresponding emission probability for blank token at $u$ and $t$  (horizontal movement in the lattice from step $t$ to $t+1$ in~\autoref{fig:lattice}).
The corresponding backward variable $\beta(t,u)$ is the probability of the sequence $\mathbf{y}_{u+1:U}$ from time $t$ to the end of the utterance.
\begin{eqnarray}
   \beta(t,u) & = & \beta(t+1, u)\varnothing(t, u) \nonumber\\
   &  & + \beta(t,u+1)y(t,u) \label{equ:backward}
\end{eqnarray}
where $\beta(T,U)=\varnothing(T,U)$.
The overall probability of sequence $y_{0:u}$ at $x_{1:T}$ is $Pr(y|x) = \alpha(T,U)\varnothing(T,U)$. 

%
%

\section{Methods}

\label{sec:methods}
Inspired by \cite{Liang2021RDropRD}, we duplicate the training data in every mini-batch to create two different views, and consistency
regularization is leveraged to boost the transducer-based system.  
The data difference comes from two folds.
First, spec augmentation~\cite{park2019specaugment} is applied to the input audio features with different maskings. Second, dropout operations~\cite{Srivastava2014DropoutAS} are utilized during model training to obtain two sets of probability distributions. The whole framework is illustrated by~\autoref{fig:tcr}. Kullbach Leibler (KL) divergence is adopted to minimize the probability distribution difference among two data sets. The overall optimization criteria is   
\begin{eqnarray}
\mathcal{L} \!= \!-  \sum_{(\mathbf{x},\mathbf{y})}\sum_{i\in[1,2]} \!\ln Pr(\mathbf{y}|\mathbf{x}^{i}) \!
 -\! \lambda D_{c}(\mathbf{x}^j|\mathbf{x}^i, \mathbf{y}) \label{equ:tot_loss}
\end{eqnarray}
where $\mathbf{x}^i$ stand for one of two different views of the training data $\mathbf{x}$ and $j=3-i \in [1,2]$ 
is the index for the other data view. $D_c$ is consistency regularization applied and $\lambda$ is the hyperparameter to integrate consistency regularization. 

A straightforward way to calculate consistency regularization is to apply KL divergence to the corresponding emission probability distributions $Pr(:|t,u)$ 
for two data views, and there are $T\times (U+1)$ emission probability distributions from the joint. 
However, we find that this simple strategy does not work in our experiments. We hypothesize that not all emission probability distributions from the joint are equal. 

Hence emission probability distributions from those points are less relevant to the optimization and could be noisy.
Applying consistency regularization to those noisy distributions could distract our trainer and hurt performance. 

Based on this hypothesis, we propose weight distributions based on their occupation probabilities and those distributions with high occupation probabilities will contribute more to the consistency regularization.


\begin{figure}
    \centering
    \includegraphics[width=1\linewidth, scale=1]{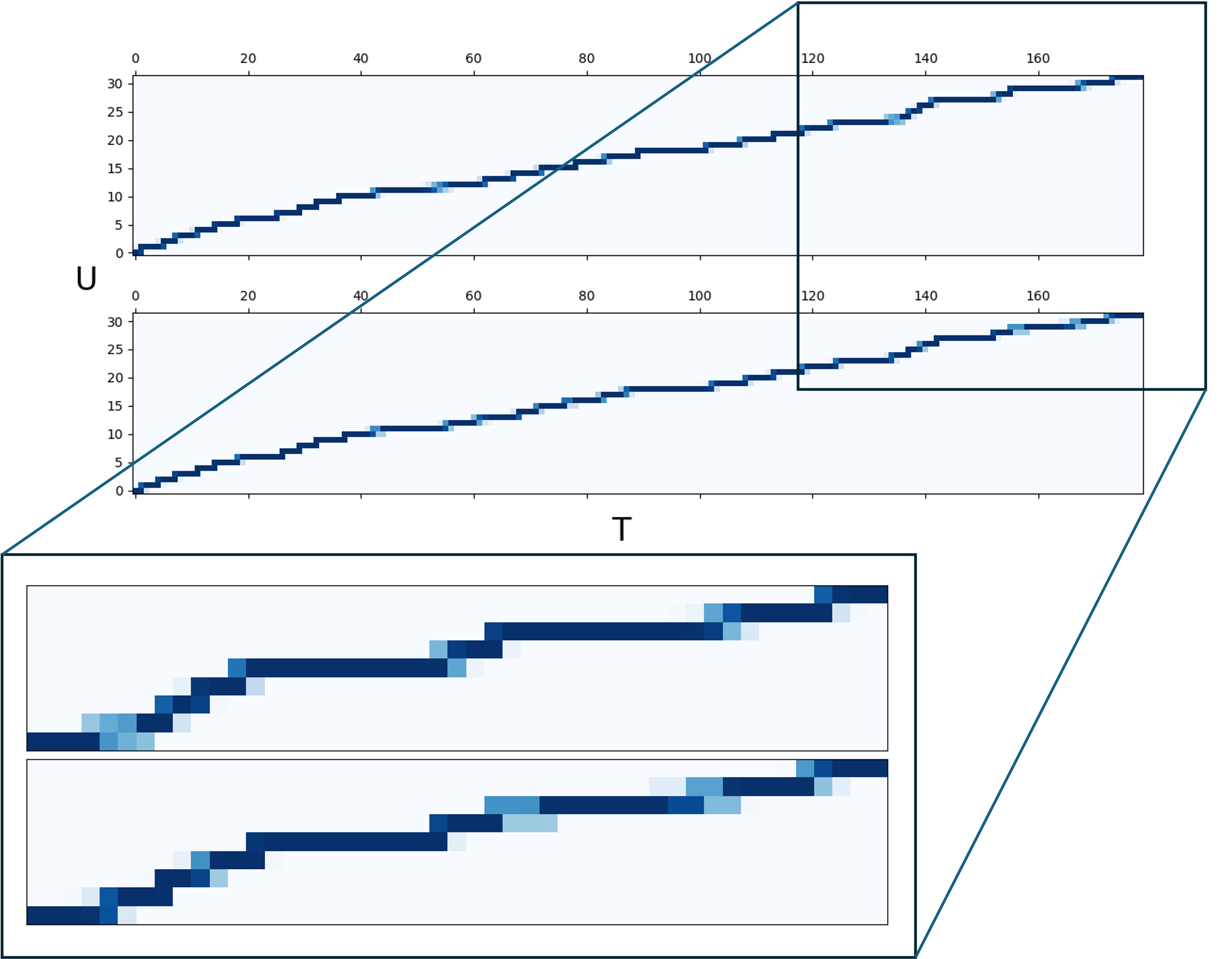}
        \caption{Occupational probability of the same audio with two different views. The occupational probability is calculated using the forward variable $\alpha$ and backward variable $\beta$. The more concentrated the blue color, the higher the occupational probability. Only a few paths have non-zero occupational probability. The two paths of different data views mostly agree except on the horizontal and vertical transitions.}
    \label{fig:align_different_views}
\end{figure}

The output sequence probability at step $t$ and token $u$ is 
\begin{eqnarray}
    Pr(t,u) & =  &\alpha(t,u)\beta(t,u) \nonumber\\
            & = & \alpha(t,u)\beta(t+1, u)\varnothing(t,u) + \nonumber\\
            && \alpha(t,u)\beta(t,u+1)y(t,u)
\end{eqnarray}
Following~\cite{graves2012sequence}, we define the total sequence probability by the sum of sequence probabilities in all points in the output lattice
\begin{equation}
    Pr(\mathbf{y}^*|\mathbf{x}) = \sum_{1 \le t+u \le T+U}\alpha(t,u) \beta(t,u)
\end{equation}

So the normalized occupation probability that the decoder might move vertical (generating non-blank token) or horizontal (generating blank token) at cell ($t$, $u$) in~\autoref{fig:lattice} are
\begin{eqnarray}
     \bar{\omega}_\varnothing(t,u)=  \frac{\alpha(t,u)\beta(t,u+1)y(t,u)}{Pr(\mathbf{y}^*|\mathbf{x})} \\
    \omega_\varnothing(t,u) =  \frac{\alpha(t,u)\beta(t+1,u)\varnothing(t,u)}{Pr(\mathbf{y}^*|\mathbf{x})} 
\end{eqnarray}

We assign different weights for different distributions according to their importance for optimization. 
The weighted consistency regularization term in~\autoref{equ:tot_loss} is defined as
\begin{eqnarray}
    && D_c(\mathbf{x}^i|\mathbf{x}^j, \mathbf{y})  =  \qquad\qquad\qquad\qquad \nonumber\\
    && \quad \beta_{\bar{\varnothing}}\frac{\sum_{t,u} \bar{\omega}_\varnothing(t,u)  D_{KL}(y_{u+1}, t,u, \mathbf{x}^i, \mathbf{x}^j)}{\sum_{t,u}\bar{\omega}_\varnothing(t,u)} + \nonumber \\
     && \quad\beta_{\varnothing}\frac{\sum_{t,u}\omega_\varnothing(t,u) D_{KL}(y_{u}, t,u, \mathbf{x}^i, \mathbf{x}^j)}{\sum_{t,u}\omega_\varnothing(t,u)},
     \label{equ:dist_weight} 
\end{eqnarray} 
where $\beta_{\bar{\varnothing}}$ and $\beta_{\varnothing}$ are weights for the non-blank token and blank token consistency regularization, and we choose both of them to 1.0 in our experiments; 
$D_{KL}(y_{u+1}, t,u, \mathbf{x}^i, \mathbf{x}^j)$ (or $D_{KL}(y_{u}, t,u, \mathbf{x}^i, \mathbf{x}^j)$) is KL divergence between the distributions $y(t,u)$ (or $\varnothing(t,u)$) from the data view $i$ and $j$.  

\autoref{fig:align_different_views} shows the occupational probability of decoding the same audio with two different views given the transcription. The different views are created by applying random spec augmentation and dropout. The top figure is the decoding of one data view, $x_i$, and the bottom figure is the decoding of another data view $x_j$. A region of the figure is zoomed-in to show a more close-up view. The two data views still result in similar diagonal paths from $(0, 0)$ to $(T, U)$. Most decoding paths will be along the diagonal line from $(0, 0)$ to $(T, U)$. Other paths that deviate from the diagonal are highly implausible. For example, paths are very unlikely to pass through the grid point $(T,0)$ or $(1, U)$ in the ASR application if the utterance is not extremely short or has zero text transcription. This observation inspired us to use the weighted sum to emphasize the regions with the highest occupational probability mass. 

In \autoref{fig:align_different_views}, the two figures mostly differ in the horizontal and vertical transitions, with blurred blue color representing multiple probable paths. In horizontal paths the occupational probability is dominated by the blank token occupational probability $\bar{\omega}_\varnothing(t,u)$, in vertical transitions it is dominated by $\omega_\varnothing(t,u)$ non blank token occupational probability. These insights give us inspiration to use occupational probability and perform a weighted sum on the divergence loss both ways so that the two views can learn from each other to have a better common representation. 

In our experiments we use the pruned region calculated by fast\_rnnt~\cite{kuang2022pruned}. (B, T, $U_r$, D), where $U_r<< U$, and $U_r$ is the number of non-blank tokens to be included in each time step $t$, to remove most of the implausible output distributions from the transducer output. 


\section{Experiment Settings}
In this work, we examine the proposed method on two different datasets for two different tasks: ASR and ST. In ASR, the input features and output tokens are monotonically aligned, while ST is a non-monotonically aligned task. 

\subsection{Data}

\textbf{ASR} We use ASR \textsc{Librispeech}~\cite{panayotov2015librispeech} dataset that contains 960 hours of data. The vocabulary is with 1024 subword units learned by SentencePiece \cite{kudo2018sentencepiece}. We evaluate the model on the dev set and report test set word error rate (WER) performance. For the baseline, we duplicate the training data without adding the TCR loss to the final loss. For TCR loss we do not duplicate the data. Thus, the total effective steps between baseline and TCR model are the same. 
\\

\noindent \textbf{ST} We use two language pairs from \textsc{Must-C} \cite{di2019must} : English to German (EN$\rightarrow$DE) and English to Spanish (EN$\rightarrow$ES), for the ST evaluation. The training data is augmented with back translation data generated from our machine translation (MT) models. The vocabulary is 1024 case-sensitive subword units learned by SentencePiece \cite{kudo2018sentencepiece}. We report the case-sensitive detokenized BLEU score. During the evaluation, we set the blank penalty~\cite{Tang2023HybridTA} to 1.0.

\begin{table}[t]
\centering
\begin{tabular}{l | c | c}
\hline
\multirow{2}{*}{Method/Dataset}   & \multicolumn{2}{c}{\textbf{test}} \\
 & \textbf{clean} & \textbf{other}\\
\hline
\hline
Conformer-L Pruned Transducer\cite{yao2023zipformer}  & 2.46& 5.55\\
Conformer-L Pruned Transducer  & 2.37 & 5.41 \\ 
\hline
Full Joint\cite{Liang2021RDropRD} &2.51&5.78 \\
Encoder MSE\cite{sapru2022using} & 2.34& 5.48 \\
Compressed Probability\cite{yu2020dual} &2.36&5.33 \\
\hline
Best One & 2.31 & 5.27 \\
Threshold  & 2.3 & 5.34 \\
TCR  & \textbf{2.28} & \textbf{5.23} \\ 
\hline
\end{tabular}
\caption{WER(\%) comparison between different consistency regularization methods on \textsc{LibriSpeech} dataset. Results are calculated using beam size of 4. The first two rows are conformer transducer baselines; The first row is result from \cite{yao2023zipformer}, the second row is ours. Full joint, encoder MSE, and compressed probability are relevant consistency regularization methods used on models trained with cross entropy loss. Best one, threshold, and TCR are our consistency regularization methods for transducer models, with TCR showing the biggest performance gain
}
\label{tab:librispeech_results}
\end{table}

\subsection{Model Configuration} 
Input speech is represented as 80-dimensional log-mel filter bank coefficients extracted every 10ms with a 25ms Hanning window. The spec augmentation~\cite{park2019specaugment} setting is with time mask 10, frequency mask 2, frequency width 27, and time width 0.05.  Utterance based channel mean normalization is applied on the input features. The subsampling factor is four.

We use Conformer transducer~\cite{gulati2020conformer} as our ASR and ST baseline. It contains 17 layers of conformer encoder, with encoder dimension 512 and 8 attention heads. The predictor is transformer based with 2 layers. The encoder and predictor have 512 hidden states while the joiner has 1025 hidden states.  The convolutional kernel size in the conformer layer is 32. We use weight decay of 1e-3. InterCTC\cite{lee2021intermediate} of 0.033 is applied at layer 5, 9, and 13. Fast\_rnnt \cite{kuang2022pruned} is used for optimization for efficient memory usage. In fast\_rnnt, lm\_scale and am\_scale are both set to 0, simple\_loss and s\_range are equal to 0.5 and 5 respectively. 

We clamp the maximum TCR loss to prevent outliers from derailing the gradient. The clamp values are 5e-3 and 0.1 for \textsc{Librispeech} and \textsc{Must-C} respectively. 
We set both $\beta_{\bar{\varnothing}}$ and $\beta_{\varnothing}$ to 1. $\lambda$ is 0.1 for \textsc{Librispeech} and 1.0 for \textsc{Must-C}.
For ST tasks, we used an encoder pre-trained on ASR task with English portion of data from the EN$\rightarrow$ES direction. We train the model with 150 epochs on 8 A100 GPUs. Each minibatch contains maximum 150 ms of audio for \textsc{Must-C} dataset and 240 ms for \textsc{Librispeech} dataset. The AdamW~\cite{loshchilov2017decoupled} optimizer was used with initial learning rate of 1.0. Learning rate warm up peaks to 10K steps then we use Noam~\cite{vaswani2017attention} for learning rate decay. Drop out rates are set to 0.1. The models are developed using the NeMo~\cite{kuchaiev2019nemo} framework. The best 5 checkpoints are averaged for inference. All of our numbers are reported without an external language model.  

\section{Results}\label{sec:results}

\subsection{Main Results}
\autoref{tab:librispeech_results} shows the ASR word error rate on the \textsc{Librispeech} test dataset with beam size of 4. The first row lists the conformer transducer baseline result from \cite{yao2023zipformer}, the second row shows our baseline result. The final row shows our proposed TCR method which improves the recognition performance compared to other methods. The model with TCR reduces 0.09 WER for test\_clean, 0.18 WER for test\_others, or 3.8\% relative WER reduction in test\_clean and 3.33\% for test\_others. We also evaluated our model trained with \textsc{Librispeech} data on internal noisy data with 2500 utterances. The baseline model achieves 32. 15\% WER while the model with TCR reduced WER to 31.42\%. TCR shows a consistent gain even on noisy data.

\subsection{Comparison of Different Consistency Regularization Implementations}

We conduct several comparison studies with different consistency regularization implementations. Encoder MSE is a method proposed in \cite{sapru2022using} which uses only encoder output trained with cross entropy loss for consistency regularization. Full joint uses the full transducer decoder output, similar to method proposed in \cite{Liang2021RDropRD} for encoder decoder model. We also studied knowledge distillation methods for transducer models\cite{yu2020dual} and apply it here for consistency regularization: Compressed Probability. Finally, we list two alternative methods alongside our main proposed TCR method: Best one path and Selection with Threshold. The results are shown in~\autoref{tab:librispeech_results}. Our proposed method (TCR) using occupational probability to perform a weighted sum on KL divergence loss gives the best WER among all implementations with one pass training. 
\\

\noindent \textbf{Full Joint}
We use the entire pruned region of the joint output and apply KL divergence. It is a naive implementation of consistency regularization considering all joint outputs equally. The full joint method produces a higher WER than that of the baseline as shown in \autoref{tab:librispeech_results}. This indicates that treating all paths equally would be detrimental to the outcome, and the method used in \cite{Liang2021RDropRD} \cite{gao2023empirical} on model trained with cross entropy loss is not applicable to model trained with transducer loss. 
\\

\noindent \textbf{Encoder MSE}
\cite{sapru2022using} uses encoder output for consistency regularization in two stage training. Table \autoref{tab:librispeech_results} encoder MSE row shows that using encoder output alone in one pass training is not effective in capturing a good representation to improve the model. 
\\

\noindent \textbf{Compressed Probability}
\label{subsec:comp_prob}
\cite{yu2020dual} and \cite{panchapagesan2021efficient} have shown good results using compressed probability for knowledge distillation. It compresses the vocabulary space into three categories, target token $y$, blank, and others. The selected region for regular consistency thus becomes (B, T, $U_r$, 3). The result is shown in row compressed probability ~\autoref{tab:librispeech_results}.  It is clear that compressing the probability within the pruned region does improve the result compared to the baseline. 
\\

\noindent \textbf{Best One Path}
In best one path implementation, we first identify the path with maximum probability in the joint output lattice. We believe that the best one path should contain the most useful information for the model to learn. However, identifying the global optimum is expensive during training. We have to first run training data in inference mode to generate best one path without any noise; Then we use the best one path as part of our training data and train a model from scratch. See \autoref{tab:librispeech_results} best one row for result. 
\\

\noindent \textbf{Selection with Threshold}
TCR gives a higher weight to paths that are close to the oracle path and less weight to bad paths that are far. In selection based regularization we have a hard cut off. Any region with emission probability below a certain threshold will not contribute to the total loss. 
We choose the top-K distributions at each step $t$ for the consistency regularization. Due to the gap between the highest probabilities between the non-blank token and blank token, we have different selections for blank and non-blank emission probability distributions. In our experiments, we select 2 best non-blank token distributions and 2 best blank token distribution at each step $t$.

The result for the selection regularization based method is shown in \autoref{tab:librispeech_results} threshold row. The results show slight improvement over baseline. 





\begin{table}
\centering
\begin{tabular}{l|c|c}
\hline
\multirow{2}{*}{Method/Dataset}  & \multicolumn{2}{c}{\textbf{tst-Common}} \\
\cline{2-3}
 & \textbf{EN$\rightarrow$DE} & \textbf{EN$\rightarrow$ES}\\
\hline
\hline
Baseline & 24.5 &  27.6 \\
\hline
TCR & 24.6 & 27.7   \\
\hline
\end{tabular}
\caption{TCR ST greedy decoding BLEU score on the \textsc{Must-C} Dataset.} 
\label{tab:mustc_result}
\end{table}

\subsection{Study on Non-Monotonic Aligned Task}
The proposed TCR relies on alignment quality and plausible paths are assigned higher weights. Compared with ASR tasks, in which the input features and output tokens are aligned monotonically, ST tasks have to deal with non-monotonic alignment. It would be challenging for the proposed method.
In \autoref{tab:mustc_result}, we compare the ST result on \textsc{Must-C} dataset with and without TCR. It shows TCR doesn't improve or degrade the performance. BLEU score is improved by 0.2 for EN$\rightarrow$DE and 0.1 for EN$\rightarrow$ES. It indicates that the proposed method is robust for non-monotonic alignment task. 


\section{Conclusion}
\label{sec:conclusion}
We introduce an effective consistency regularization method for the transducer model. It uses occupational probability to give emphasis to lattice output with good alignments. The model learns to minimize the different distributions introduced by data spec augmentation and drop out. We show that our method is beneficial particularly on ASR applications. Further work would be to improve more on non-monotonic tasks such as speech translation. 

\bibliographystyle{IEEEbib}
\bibliography{refs}
\label{sec:ref}

\end{document}